%% file: main.tex
\documentclass[letterpaper, 10 pt, conference]{ieeeconf}
\IEEEoverridecommandlockouts
\usepackage[T1]{fontenc}
\usepackage{times}

\usepackage{cite}
\usepackage{amsmath,amssymb,amsfonts}
\usepackage{bm}
\usepackage{graphicx}
\usepackage{xcolor}
\usepackage{booktabs}
\usepackage{multirow}
\usepackage{array}
\usepackage{caption}
\usepackage{subcaption}
\usepackage{algorithm}
\usepackage[noend]{algpseudocode}
\usepackage{tikz}
\usetikzlibrary{positioning,arrows.meta,calc,fit,backgrounds,shapes.geometric}
\usepackage{flushend}
\usepackage[colorlinks]{hyperref}
\hypersetup{colorlinks, linkcolor=black, citecolor=black, urlcolor=cyan}

\input{macros.tex}

\title{\LARGE \bf
\ours: Few-Step Flow Matching Policies for\\ Distilling Multi-Gait Humanoid Locomotion}

\author{
Savio Williams,
Siddhanth Ramesh,
and Daan Visser\\
University of Twente, Enschede, The Netherlands
}

\begin{document}
\makeatletter
\providecommand{\bstctlcite}[1]{\@bsphack\@for\@citeb:=#1\do{\edef\@citeb{\expandafter\@firstofone\@citeb}\if@filesw\immediate\write\@auxout{\string\citation{\@citeb}}\fi}\@esphack}
\makeatother
\bstctlcite{IEEEexample:BSTcontrol}

\maketitle
\thispagestyle{empty}
\pagestyle{empty}

\input{sections/abstract}
\input{sections/introduction}
\input{sections/related}
\input{sections/preliminaries}
\input{sections/method}
\input{sections/experiments}
\input{sections/results}
\input{sections/conclusion}

\bibliographystyle{IEEEtran}
\bibliography{references}

\end{document}

%% file: macros.tex
\newif\ifdraft
\draftfalse
\ifdraft

\else

\fi

\newcommand{\ours}{FlowStride}
\newcommand{\birodiff}{BiRoDiff}

\newcommand{\norm}[1]{\left\lVert#1\right\rVert}
\newcommand{\R}{\mathbb{R}}
\newcommand{\E}{\mathbb{E}}
\newcommand{\N}{\mathcal{N}}
\newcommand{\D}{\mathcal{D}}
\newcommand{\M}{\mathcal{M}_\phi}
\newcommand{\vt}{v_\theta}
\newcommand{\Ob}{\mathbf{O}}
\newcommand{\A}{\mathbf{A}}
\newcommand{\Ahat}{\hat{\mathbf{A}}}
\newcommand{\Atil}{\tilde{\mathbf{A}}}
\newcommand{\ba}{\mathbf{a}}
\newcommand{\bs}{\mathbf{s}}
\newcommand{\bz}{\mathbf{z}}
\newcommand{\beps}{\boldsymbol{\epsilon}}
\newcommand{\shift}{\mathcal{S}}
\newcommand{\Wtwo}{W_2}
\newcommand{\tr}{\operatorname{tr}}
\newcommand{\Cov}{\operatorname{Cov}}

\newtheorem{proposition}{Proposition}
\newtheorem{assumption}{Assumption}
\newtheorem{remark}{Remark}

\newenvironment{pf}{\noindent\textit{Proof:}\ }{\hfill$\blacksquare$\par\smallskip}

%% file: sections/abstract.tex
\begin{abstract}
Humanoid robots often come with several locomotion controllers, and distilling them offline into a single policy is attractive. The pooled data are multimodal, however, because different experts respond to the same observation with different gaits. Diffusion policies can represent this multimodality, but they need tens of sequential network evaluations per control step, and their stochastic sampling can switch gaits from one step to the next. We present \ours{}, a conditional flow-matching policy for bipedal locomotion that maps a proprioceptive history to short chunks of joint-position targets with a few Euler steps of a deterministic ODE. Its velocity field is parameterised through a denoiser, and a warm-started receding-horizon variant reuses the shifted previous chunk. We prove three properties. A single cold-start Euler step collapses to behaviour cloning. With fixed noise the policy is a Lipschitz feedback law, whereas fresh-noise samplers have an action-jitter floor. The distributional error of warm-starting is bounded in Wasserstein distance. We evaluate on a multi-gait distillation task on the Unitree G1 in MuJoCo, in which a fast and a slow gait expert are pooled into $500{,}000$ offline samples without the signal that distinguishes them. With four network evaluations ($14.6\times$ fewer floating-point operations than a $60$-step DDPM policy), \ours{} walks in all flat-ground episodes, including at held-out and extrapolated speeds. It tracks velocity best among the learned policies, reproduces both gaits, and switches gait $3.1\times$ less often within an episode than DDPM. Behaviour cloning falls in $18\%$ of the episodes and collapses mostly onto one gait. Under severe unseen perturbations, all distilled policies fail where the experts partially succeed, a limitation we analyse.
\end{abstract}

%% file: sections/introduction.tex
\section{Introduction}
\label{sec:intro}

Humanoid robots such as the Unitree G1 and H1 are increasingly equipped with learned locomotion controllers~\cite{radosavovic2024real,li2024versatile,gu2024humanoidgym,zakka2025playground,unitree2023rlgym} and with model-based ones~\cite{dicarlo2018convex,kajita2001lip,pratt2006capture}. Practitioners therefore accumulate several controllers for the same robot, with different gaits, speeds or styles. Learning a single controller \emph{offline} from such a pool is attractive. It reuses existing behaviours without further environment interaction, it can combine their strengths, and it replaces a collection of controllers by one network with a fixed inference cost.

\begin{figure}[t]
    \centering
    \resizebox{\linewidth}{!}{\input{figures/overview.tex}}
    \caption{\textbf{Overview.} Two gait experts, obtained by driving the gait clock of a pretrained G1 locomotion policy at stride periods of $0.6$\,s and $1.0$\,s, are recorded and pooled into one offline dataset without the clock signal. The conditional action distribution of such pooled data can be multimodal (inset, illustrative): behaviour cloning regresses to the mean of the two gaits, whereas generative policies commit to one. \ours{} needs only a few network evaluations per control step.}
    \label{fig:overview}
\end{figure}
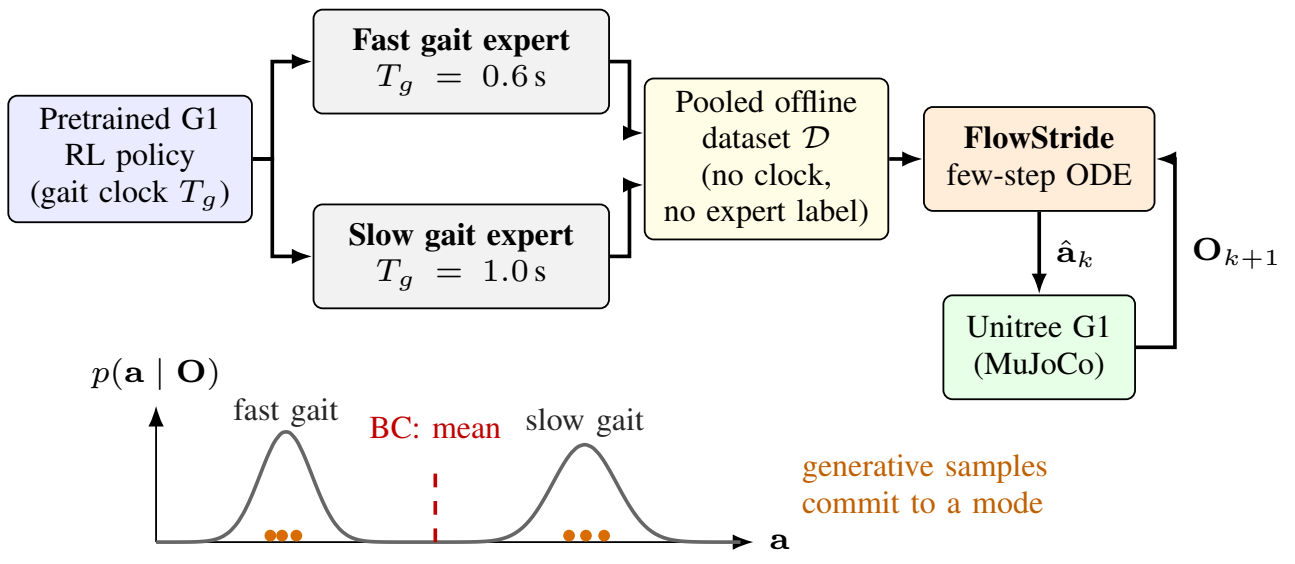

The difficulty is that pooled data are \emph{multimodal}. Two gait experts commanded to walk at the same speed take strides of different length and cadence. A proprioceptive observation can therefore be followed by different actions depending on which gait is being executed. Regressing actions with a mean-squared error averages incompatible gaits~\cite{mandlekar2021matters} (Fig.~\ref{fig:overview}). Generative policies avoid this averaging. Diffusion models~\cite{ho2020denoising,sohl2015deep} have been used for planning~\cite{janner2022planning,ajay2022conditional}, visuomotor manipulation~\cite{chi2023diffusion} and quadrupedal locomotion~\cite{huang2024diffuseloco}. \birodiff{}~\cite{mothish2024birodiff} showed that a lightweight DDPM policy for a six-degree-of-freedom point-foot biped can capture several speeds and terrains in one controller and walk on terrain absent from its training data.

Scaling this approach to humanoids exposes three limitations of diffusion policies.
\begin{enumerate}
    \item \emph{Inference cost.} Each action of \birodiff{} requires $60$ sequential network evaluations. With an action head sized for a humanoid, this amounts to $206$\,MFLOP per control step, which strains on-board computers.
    \item \emph{Myopia.} Predicting only the next action makes the policy short-sighted~\cite{mothish2024birodiff}, and predicting longer horizons multiplies the cost unless sampling becomes cheap.
    \item \emph{Stochastic step-to-step sampling.} Ancestral DDPM sampling draws fresh noise at every denoising step and every control step, so consecutive actions are independent draws from the conditional distribution. When its modes are different gaits, the policy can switch gaits mid-stride.
\end{enumerate}

\emph{Flow matching}~\cite{lipman2023flow,liu2023rectified,albergo2023building} offers a direct remedy. It trains a velocity field by simulation-free regression along simple, typically straight, paths between noise and data, and samples by integrating a deterministic ODE. Straight paths can be integrated accurately with very few steps, and a deterministic sampler turns the generative policy into an ordinary feedback law. Flow-matching policies have been adopted for manipulation and vision-language-action models~\cite{black2024pi0,braun2024riemannian,zhang2024affordance,chisari2024pointflowmatch,zhang2025flowpolicy}. In legged locomotion, flow policies have so far been trained mainly with online RL~\cite{mcallister2025fpo,zhang2025reinflow,yi2026fpopp,huang2026rfpo}, distilled with online expert queries~\cite{wang2026omnixtreme}, or bootstrapped from sampling-based predictive control~\cite{kurtz2026gpc}. Purely offline distillation of several gait experts into one real-time generative policy for a humanoid has, to our knowledge, not been studied in a controlled setting.

We introduce \ours{} (Fig.~\ref{fig:overview}), a flow-matching policy for bipedal humanoid locomotion. We evaluate it on a multi-gait distillation task on the Unitree G1 in MuJoCo, in which a fast and a slow gait expert are pooled without the signal that distinguishes them. Our contributions are:
\begin{enumerate}
    \item \textbf{A conditional flow-matching locomotion policy} that maps a proprioceptive history to chunks of joint-position targets with one to four ODE steps. Its velocity field is parameterised through a denoiser, which our analysis shows to be exact and which trains far faster than the plain velocity target.
    \item \textbf{Warm-started receding-horizon chunking.} Following warm-starting schemes for generative policies~\cite{kurtz2026gpc,hoeg2025sdp}, each integration starts from the noised, time-shifted previous chunk. By construction, this reduces the steady-state cost to two network evaluations per control step, $27\times$ fewer floating-point operations than an equally sized diffusion policy. Its closed-loop evaluation is left to future work.
    \item \textbf{Analysis.} One cold-start Euler step is exactly behaviour cloning, whereas one warm-started step returns a posterior mean anchored to the previous plan. For a fixed noise sample the policy is Lipschitz in its observations, in contrast to the jitter floor $2\tr\Cov[\ba\mid\Ob]$ of samplers that draw fresh noise at each step. We also bound the Wasserstein error of warm-starting.
    \item \textbf{A controlled evaluation} against both experts, behaviour cloning, a diffusion policy and few-step deterministic diffusion sampling. It covers seen, held-out and extrapolated commands, six unseen terrain and perturbation conditions, and metrics for gait-mode fidelity.
\end{enumerate}
With four network evaluations per step, \ours{} matches the $60$-step DDPM policy in reliability on flat ground at $14.6\times$ lower cost. It tracks velocity best among the learned policies, and it switches gait within an episode $3.1\times$ less often than DDPM. Behaviour cloning, in contrast, falls in $18\%$ of the episodes and collapses mostly onto one gait. Under severe unseen perturbations, all distilled policies fail where the experts partially succeed, which identifies data coverage as the main limitation of offline distillation.

The remainder of the paper is organised as follows. Section~\ref{sec:related} reviews related work, and Section~\ref{sec:prelim} introduces the platform, the experts and flow matching. Section~\ref{sec:method} presents \ours{} and its analysis. Sections~\ref{sec:experiments} and~\ref{sec:results} describe the experimental protocol and results, and Section~\ref{sec:conclusion} concludes.

%% file: figures/overview.tex
\begin{tikzpicture}[
  font=\scriptsize,
  >={Latex[length=1.6mm]},
  box/.style={draw, rounded corners=2pt, align=center, inner sep=2.5pt, minimum height=8mm},
  exp/.style={box, fill=gray!10, text width=21mm},
  src/.style={box, fill=blue!8, text width=17mm},
  dat/.style={box, fill=yellow!12, text width=17mm},
  pol/.style={box, fill=orange!15, text width=16mm},
  rob/.style={box, fill=green!10, text width=13mm},
  arr/.style={->, thick},
]
\node[src] (pre) at (0,0) {Pretrained G1\\ RL policy\\ (gait clock $T_g$)};
\node[exp] (fast) at (2.55,0.75) {\textbf{Fast gait expert}\\ $T_g=0.6$\,s};
\node[exp] (slow) at (2.55,-0.75) {\textbf{Slow gait expert}\\ $T_g=1.0$\,s};
\node[dat] (data) at (4.9,0) {Pooled offline\\ dataset $\D$\\ (no clock,\\ no expert label)};
\node[pol] (flow) at (7.0,0) {\textbf{\ours{}}\\ few-step ODE};
\node[rob] (g1) at (7.0,-1.45) {Unitree G1\\ (MuJoCo)};

\draw[arr] (pre.east) -- ++(0.15,0) |- (fast.west);
\draw[arr] (pre.east) -- ++(0.15,0) |- (slow.west);
\draw[arr] (fast.east) -- ++(0.15,0) |- ([yshift=2mm]data.west);
\draw[arr] (slow.east) -- ++(0.15,0) |- ([yshift=-2mm]data.west);
\draw[arr] (data) -- (flow);
\draw[arr] (flow) -- node[right]{$\hat\ba_k$} (g1);
\draw[arr] (g1.east) -- ++(0.3,0) |- node[pos=0.25, right]{$\Ob_{k+1}$} (flow.east);

\begin{scope}[shift={(0.2,-2.95)}]
  \draw[->] (0,0) -- (4.6,0) node[right]{$\ba$};
  \draw[->] (0,0) -- (0,1.05) node[above]{$p(\ba\mid\Ob)$};
  \draw[thick, gray!80!black, domain=0:4.5, samples=80, smooth]
    plot (\x, {0.85*exp(-((\x-1.0)^2)/0.08) + 0.75*exp(-((\x-3.3)^2)/0.12)});
  \node[gray!30!black] at (1.0,1.0) {fast gait};
  \node[gray!30!black] at (3.3,0.92) {slow gait};
  \draw[red!75!black, dashed, thick] (2.15,0) -- (2.15,0.62);
  \node[red!75!black, align=center] at (2.15,0.88) {BC: mean};
  \foreach \x in {0.88,0.97,1.08,3.18,3.31,3.45}
    \fill[orange!85!black] (\x,0.05) circle (0.045);
  \node[orange!75!black, anchor=west, align=left] at (4.85,0.45) {generative samples\\ commit to a mode};
\end{scope}
\end{tikzpicture}

%% file: sections/related.tex
\section{Related Work}
\label{sec:related}

\subsection{Bipedal and Humanoid Locomotion Control}
Model-based bipedal control builds on reduced-order models such as the linear inverted pendulum~\cite{kajita2001lip} and the capture point~\cite{pratt2006capture}. Convex MPC over single-rigid-body dynamics~\cite{dicarlo2018convex} combined with whole-body control is a standard recipe for legged robots. Model-free RL has produced robust blind controllers for bipeds~\cite{siekmann2021blind,li2024versatile} and humanoids~\cite{radosavovic2024real,gu2024humanoidgym}, often trained with massively parallel simulation~\cite{rudin2022learning,zakka2025playground}. Reachability analysis has been used to make such learned locomotion policies safe~\cite{rao2026reachability}. Distilling optimisation-based controllers into neural policies dates back to guided policy search~\cite{levine2013gps} and MPC-Net~\cite{carius2020mpcnet}, both of which query the optimiser during training. We study the purely offline setting~\cite{levine2020offline}, in which a fixed pool of trajectories from several gait experts is distilled into a single generative policy without rewards, interaction or expert queries.

\subsection{Diffusion Policies for Control and Locomotion}
Diffusion models have been used as trajectory planners~\cite{janner2022planning,ajay2022conditional}, as visuomotor policies that predict receding-horizon action chunks~\cite{chi2023diffusion}, and as expressive policy classes for offline RL~\cite{wang2023diffusionql}. For legged robots, DiffuseLoco~\cite{huang2024diffuseloco} learns a diffusion policy from offline multi-skill datasets and deploys it on real quadrupeds, and guided diffusion has been used for versatile humanoid control~\cite{liao2025beyondmimic}. \birodiff{}~\cite{mothish2024birodiff}, on which this work builds, is an MLP-based DDPM policy that captures two speeds and two terrain types in one controller for a point-foot biped and generalises to unseen terrain. A recurring obstacle is the cost of iterative denoising. DDIM~\cite{song2021ddim} reduces the number of steps, and consistency models~\cite{song2023consistency} reach one or two steps at the price of an additional distillation stage. \ours{} obtains few-step sampling directly from its training objective.

\subsection{Flow Matching Policies}
Flow matching~\cite{lipman2023flow,liu2023rectified,albergo2023building} replaces the noising chain of diffusion with regression onto a velocity field along simple interpolation paths. In robotics it underlies the action expert of the $\pi_0$ vision-language-action model~\cite{black2024pi0}, Riemannian, point-cloud and affordance-conditioned manipulation policies~\cite{braun2024riemannian,chisari2024pointflowmatch,zhang2024affordance}, multi-support humanoid manipulation~\cite{rouxel2024flow}, and consistency-trained one-step policies~\cite{zhang2025flowpolicy}. Strictly one-step generators can also be obtained with shortcut models~\cite{frans2025shortcut} or mean flows~\cite{geng2025meanflow}. Flow policies can be trained or fine-tuned with RL~\cite{park2025fql,mcallister2025fpo,zhang2025reinflow}, including offline RL with safety constraints via epigraph and reachability formulations~\cite{tayal2026bridging,tayal2026safeflowq}. Recent work applies such methods to legged and humanoid locomotion, including on hardware~\cite{yi2026fpopp,huang2026rfpo}, and distils humanoid motion-tracking experts into flow policies with online expert queries~\cite{wang2026omnixtreme}. Closest to our work, Generative Predictive Control (GPC)~\cite{kurtz2026gpc} iteratively trains flow-matching policies on data from sampling-based predictive control and deploys them in receding horizon with a warm start, on systems that include a planar walker. \ours{} differs from these works in three respects. It is trained purely offline on pooled data from multiple gait experts, it controls a humanoid in closed loop with a budget of a few network evaluations per step, and it is compared one-to-one with a diffusion policy that has identical data and network sizes.

\subsection{Action Chunking and Warm-Starting}
Predicting action chunks improves temporal coherence in imitation learning. ACT~\cite{zhao2023act} averages overlapping chunks with temporal ensembling, Diffusion Policy executes chunks in receding horizon~\cite{chi2023diffusion}, and bidirectional decoding selects among sampled chunks at test time~\cite{liu2025bid}. Real-time chunking~\cite{black2025rtc} executes chunks asynchronously and conditions each new chunk on the already committed prefix. Several works reuse the previous solution to reduce sampling cost. Streaming Diffusion Policy~\cite{hoeg2025sdp} rolls a partially denoised trajectory forward by one step, and GPC~\cite{kurtz2026gpc} initialises flow integration at $(1-\alpha)\beps+\alpha\bar{\mathbf{U}}_{k-1}$ from the previous solution. Streaming flow policies~\cite{jiang2025streaming} and continuation-based schemes~\cite{liu2026legato} start the flow from previously executed actions. These schemes build on partial noising as in SDEdit~\cite{meng2022sdedit}. Our warm start~\eqref{eq:warm} combines the GPC initialisation with the one-step time shift of~\cite{hoeg2025sdp}. We do not claim the mechanism as a contribution. Instead, we characterise its single-step behaviour (Proposition~\ref{prop:onestep}) and bound its distributional error (Proposition~\ref{prop:warm}).

%% file: sections/preliminaries.tex
\section{Preliminaries}
\label{sec:prelim}

\textbf{Notation.} As in~\cite{mothish2024birodiff}, subscripts index control time steps $k\in\mathbb{N}$ and superscripts index generative (flow) time $\tau\in[0,1]$, with $\tau=0$ corresponding to noise and $\tau=1$ to data. $\norm{\cdot}$ is the Euclidean norm and $\Wtwo$ the $2$-Wasserstein distance.

\subsection{Platform and Policy Interface}
\label{sec:platform}
We use the Unitree G1 humanoid in MuJoCo~\cite{todorov2012mujoco}, with the 12-DoF model released with Unitree's official locomotion code~\cite{unitree2023rlgym}. Each leg has six actuated joints (hip pitch, roll and yaw, knee, ankle pitch and roll), and the upper body is rigidly attached to the pelvis. Physics runs at $1$\,kHz. The policy runs at $f_c=50$\,Hz and outputs joint-position targets $\ba_k\in\R^{12}$, expressed as offsets from a nominal standing pose $\mathbf{q}_0$. Within each control interval, a joint-level PD law tracks these targets at the simulation rate:
\begin{equation}
    \boldsymbol{\tau}(t)=K_p\big(\mathbf{q}_0+\ba_k-\mathbf{q}(t)\big)-K_d\,\dot{\mathbf{q}}(t),
    \label{eq:pd}
\end{equation}
with the diagonal gains of the official deployment configuration ($K_p=100$, $150$ and $40$\,N\,m/rad for the hips, knees and ankles). The proprioceptive state $\bs_k\in\R^{45}$ contains the pelvis angular velocity, the projected gravity vector, the velocity command $\mathbf{c}=(v_x,v_y,\omega_z)$, the joint-position offsets $\mathbf{q}-\mathbf{q}_0$, the joint velocities, and the previous action. It deliberately excludes the base linear velocity, which is not directly measured on hardware. The policy observes a history of five states, $\Ob_k=(\bs_{k-4},\dots,\bs_k)\in\R^{225}$.

\subsection{Gait Experts}
\label{sec:experts}
Our experts are derived from Unitree's pretrained G1 locomotion policy~\cite{unitree2023rlgym}, an LSTM trained with RL. Besides proprioception and the command, it observes a periodic gait clock $(\sin 2\pi\phi,\cos 2\pi\phi)$, $\phi=(t \bmod T_g)/T_g$, that dictates the footfall timing. The released policy uses a stride period of $T_g=0.8$\,s. We found that it walks stably and tracks commands for stride periods between $0.6$ and $1.1$\,s. We therefore define two experts by driving the clock at different periods. The \emph{fast} expert uses $T_g=0.6$\,s and the \emph{slow} expert uses $T_g=1.0$\,s. At the same commanded speed they produce clearly different gaits: the fast expert takes $40\%$ shorter, higher-cadence strides. The experts' actions are already position targets of the form~\eqref{eq:pd}, so they are used directly.

\subsection{Problem Formulation}
We are given offline trajectories $\D=\D_{\text{fast}}\cup\D_{\text{slow}}$ recorded from the two experts over a range of velocity commands. No reward labels, further environment interaction or expert queries are available. The learner observes neither the gait clock nor the identity of the expert. The conditional action distribution $p(\ba_k\mid\Ob_k)$ of the pooled data is therefore a mixture over gaits, and it is multimodal wherever the observation history does not reveal which gait is being executed. The goal is a single policy that (i) reproduces both gaits without mixing them within an episode, (ii) transfers to unseen commands, terrain and perturbations, (iii) runs in closed loop at $f_c$ on modest compute, and (iv) produces smooth joint targets.

\subsection{Flow Matching}
\label{sec:prelim_fm}
Let $q$ be a data distribution on $\R^d$ and $p_0=\N(0,I)$. Flow matching~\cite{lipman2023flow,liu2023rectified,albergo2023building} learns a time-dependent velocity field $v:\R^d\times[0,1]\to\R^d$ whose flow $\frac{d}{d\tau}x^\tau=v(x^\tau,\tau)$, started at $x^0\sim p_0$, produces $x^1\sim q$. With the linear (rectified-flow / optimal-transport) interpolant
\begin{equation}
    x^\tau=(1-\tau)\,x^0+\tau\,x^1,\qquad x^0\sim p_0,\; x^1\sim q,
    \label{eq:interp}
\end{equation}
drawn independently, the conditional velocity along each path is constant, $\dot x^\tau = x^1-x^0$. The conditional flow-matching (CFM) loss
\begin{equation}
    \mathcal{L}(\theta)=\E_{\tau,x^0,x^1}\norm{\vt(x^\tau,\tau)-(x^1-x^0)}^2
    \label{eq:cfm}
\end{equation}
is minimised by the marginal velocity field
\begin{equation}
    u(x,\tau)=\E\!\left[x^1-x^0 \,\middle|\, x^\tau=x\right],
    \label{eq:marginal}
\end{equation}
which satisfies the continuity equation for the marginal path $p_\tau=\mathrm{Law}(x^\tau)$ and therefore transports $p_0$ to $q$~\cite{lipman2023flow,liu2023rectified}. Training is simulation-free regression. Sampling integrates the ODE from $\tau=0$ to $1$, typically with a handful of Euler steps, because the conditional paths are straight and the marginal paths are close to straight when $q$ is concentrated around a single mode. Conditioning on a context $\bz$ (here, the latent observation) amounts to replacing $q$ by $q(\cdot\mid\bz)$ and giving $\bz$ as an extra input to $\vt$.

By contrast, DDPM~\cite{ho2020denoising} defines a stochastic forward noising chain and samples by \emph{ancestral} sampling, which adds fresh Gaussian noise at every reverse step. Deterministic samplers such as DDIM~\cite{song2021ddim} reduce the number of steps for a trained diffusion model. They integrate the curved probability-flow ODE of the variance-preserving process~\cite{song2021score}, however, which tolerates very coarse discretisation less well than the straight paths of~\eqref{eq:interp}~\cite{liu2023rectified}.

%% file: sections/method.tex
\section{\ours{}: Flow Matching Policies for Bipedal Locomotion}
\label{sec:method}

\begin{figure*}[t]
    \centering
    \resizebox{\textwidth}{!}{\input{figures/arch.tex}}
    \caption{\textbf{\ours{} at control step $k$.} The proprioceptive history is embedded by the encoder $\M$. Starting from a warm-start point at flow time $\tau_0$, which mixes a noise sample $\beps$ with the previous chunk shifted by one step, $\shift(\Ahat_{k-1})=(\hat\ba_{k},\dots,\hat\ba_{k+H-2},\hat\ba_{k+H-2})$, the learned velocity field $\vt$ is integrated with $N(1-\tau_0)$ Euler steps of size $\Delta=1/N$. This yields an $H$-step chunk of joint-position targets, of which only the first is sent to the joint PD controllers (receding horizon). At the first step of an episode, integration starts from pure noise ($\tau_0=0$).}
    \label{fig:arch}
\end{figure*}
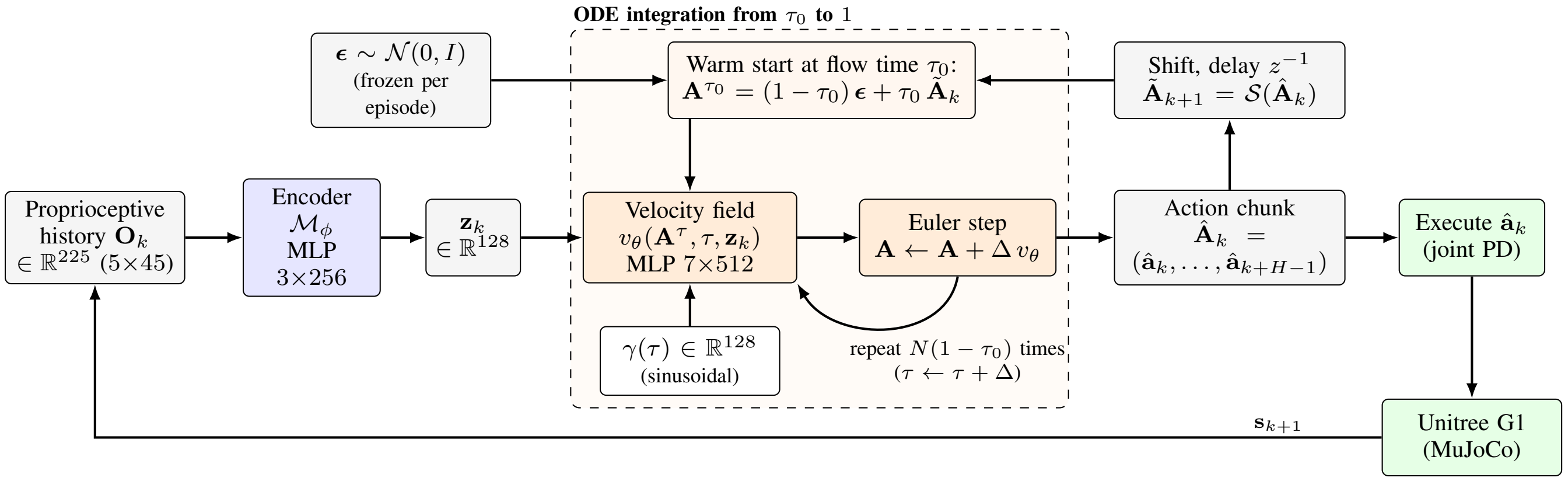

\ours{} replaces the noise-prediction network and DDPM sampler of \birodiff{} with a conditional velocity field and an ODE sampler, and extends the output from one action to a chunk of $H$ actions that is executed in a receding-horizon fashion (Fig.~\ref{fig:arch}).

\subsection{Policy Parameterisation}
\label{sec:param}
\textbf{Latent observations.} As in \birodiff{}, the observation history is embedded by a learnable encoder $\bz_k=\M(\Ob_k)\in\R^{128}$. It is an MLP with three hidden layers of width $256$ and GELU activations~\cite{gelu}, scaled up from \birodiff{} for the larger observation space. The encoder is trained end-to-end with the velocity field, so $\bz_k$ is a task-specific latent observation rather than a fixed feature.

\textbf{Action chunks.} The policy generates a chunk of $H$ future joint-position targets,
\begin{equation}
    \A_k=(\ba_k,\ba_{k+1},\dots,\ba_{k+H-1})\in\R^{12H},
\end{equation}
with each joint normalised to $[-1,1]$ using the minimum and maximum over the dataset. Training chunks are cut from the recorded episodes. Near the end of an episode they are padded by repeating the last action, as in~\cite{chi2023diffusion}. With $H=1$ the output is a single action, as in \birodiff{}.

\textbf{Velocity field.} $\vt(\A,\tau,\bz)$ maps a noisy chunk, the flow time and the latent observation to a velocity in $\R^{12H}$. Its input is the concatenation $[\A,\gamma(\tau),\bz]$, where $\gamma$ is a $128$-dimensional sinusoidal embedding of $1000\,\tau$. It is an MLP with seven hidden layers of width $512$ and GELU activations, the depth of the \birodiff{} denoiser at twice its width. Because the hidden layers dominate the cost, a chunk of $H=8$ ($0.16$\,s) increases the multiply--accumulate operations per network evaluation by only $5\%$ (Table~\ref{tab:compute}).

\subsection{Training Objective}
\label{sec:training}
Given $(\Ob,\A)\sim\D$, Gaussian noise $\beps\sim\N(0,I)$ and $\tau\sim\mathcal{U}[0,1]$, we form the interpolant $\A^\tau=(1-\tau)\beps+\tau\A$ of~\eqref{eq:interp} and minimise the conditional flow-matching loss
\begin{equation}
    \mathcal{L}(\theta,\phi)=\E\Big\|\vt\big(\A^\tau,\tau,\M(\Ob)\big)-(\A-\beps)\Big\|^2
    \label{eq:loss}
\end{equation}
jointly over the velocity-field parameters $\theta$ and the encoder parameters $\phi$. Compared with the \birodiff{} objective, only the regression target changes, from the noise $\beps$ to the velocity $\A-\beps$, and the discrete diffusion index $t\in\{1,\dots,60\}$ is replaced by a continuous $\tau\in[0,1]$. No noise schedule has to be designed. We sample $\tau$ uniformly; the velocity field is parameterised through a denoiser (Sec.~\ref{sec:experiments}).

\subsection{Few-Step ODE Sampling}
\label{sec:sampling}
At deployment, a chunk is generated by integrating $\frac{d}{d\tau}\A^\tau=\vt(\A^\tau,\tau,\bz_k)$ with the explicit Euler method on the uniform grid $\tau_n=n/N$:
\begin{equation}
    \A^{\tau_{n+1}}=\A^{\tau_n}+\tfrac{1}{N}\,\vt(\A^{\tau_n},\tau_n,\bz_k).
    \label{eq:euler}
\end{equation}
A \emph{cold start} begins from $\A^{0}=\beps$ and costs $N$ network function evaluations (NFE). Because the conditional paths~\eqref{eq:interp} are straight, small $N$ already gives accurate integration wherever the action distribution is unimodal. Section~\ref{sec:analysis} shows, however, that the extreme case $N=1$ is degenerate.

\subsection{Warm-Started Receding-Horizon Execution}
\label{sec:warm}
Executing a whole chunk open loop would sacrifice the closed-loop reactivity that a humanoid needs for balance. \ours{} therefore re-plans at every control step and executes only the first action $\hat\ba_k$ of the chunk $\Ahat_k$. Re-planning from scratch, however, discards the information in the previous chunk and costs $N$ NFE per step. We instead \emph{warm-start} the integration. Let the shift operator be
\begin{equation}
    \shift(\ba_0,\dots,\ba_{H-1})=(\ba_1,\dots,\ba_{H-1},\ba_{H-1}),
\end{equation}
which advances a plan by one control step and repeats its last element. At step $k>0$ we form the proposal $\Atil_k=\shift(\Ahat_{k-1})$ and place it on the interpolation path at a grid time $\tau_0=n_0/N$:
\begin{equation}
    \A_k^{\tau_0}=(1-\tau_0)\,\beps+\tau_0\,\Atil_k.
    \label{eq:warm}
\end{equation}
We then integrate~\eqref{eq:euler} from $\tau_0$ to $1$ with the current latent observation $\bz_k$. This uses $N(1-\tau_0)$ NFE, i.e.\ two for our default $N=4$, $\tau_0=0.5$. Eq.~\eqref{eq:warm} has exactly the form of the training interpolant, so the warm-start point lies in the region of $(\A,\tau)$-space on which $\vt$ was trained. It follows the partial-noising principle of SDEdit~\cite{meng2022sdedit}. It coincides with the warm start of GPC~\cite{kurtz2026gpc} applied to the time-shifted plan, as in Streaming Diffusion Policy~\cite{hoeg2025sdp}. The parameter $\tau_0$ trades consistency with the previous plan against freedom to re-plan: $\tau_0=0$ recovers cold-start sampling, and $\tau_0\to1$ replays the previous plan.

\textbf{Noise handling.} By default, $\beps$ is drawn once at the start of an episode and kept fixed (\emph{frozen noise}). The controller is then a deterministic dynamical system on the pair (plan, observation), and the analysis of Proposition~\ref{prop:lipschitz} applies. A \emph{stochastic} variant redraws $\beps$ at every control step and is covered by Proposition~\ref{prop:warm}. Algorithm~\ref{alg:sample} summarises the controller.

\begin{algorithm}[t]
\caption{\ours{} receding-horizon control}
\label{alg:sample}
\begin{algorithmic}[1]
\Require $\M$, $\vt$, steps $N$, warm-start time $\tau_0=n_0/N$
\State Sample $\beps\sim\N(0,I)$ \Comment{frozen for the episode}
\For{$k=0,1,2,\dots$}
    \State $\bz_k\gets\M(\Ob_k)$
    \If{$k=0$} $\A\gets\beps$, \ $n\gets0$ \Comment{cold start}
    \Else \ $\A\gets(1-\tau_0)\beps+\tau_0\,\shift(\Ahat_{k-1})$, \ $n\gets n_0$
    \EndIf
    \While{$n<N$}
        \State $\A\gets\A+\frac{1}{N}\vt(\A,n/N,\bz_k)$; \ $n\gets n+1$
    \EndWhile
    \State $\Ahat_k\gets\A$; apply $\hat\ba_k$ (first element of $\Ahat_k$) via joint PD
    \State Observe $\bs_{k+1}$ and update $\Ob_{k+1}$
\EndFor
\end{algorithmic}
\end{algorithm}

\subsection{Analysis}
\label{sec:analysis}
We now explain why one cold-start step is insufficient, why deterministic flow sampling produces smooth actions, and how much error warm-starting introduces. Propositions~\ref{prop:onestep} and~\ref{prop:warm} concern the exact minimiser $u$ of~\eqref{eq:loss}, i.e.\ the conditional marginal velocity $u(\A,\tau,\bz)=\E[\A^1-\A^0\mid\A^\tau=\A,\bz]$. Proposition~\ref{prop:lipschitz} concerns the learned field $\vt$ under the following regularity assumption.

\begin{assumption}
\label{ass:lip}
The velocity field is $L_A$-Lipschitz in $\A$ and $L_z$-Lipschitz in $\bz$, uniformly in $\tau\in[0,1]$, and the encoder $\M$ is $L_M$-Lipschitz.
\end{assumption}
Assumption~\ref{ass:lip} holds for any MLP with Lipschitz activations such as GELU. For the exact field of a multimodal distribution, $L_A$ can grow near $\tau=1$, so the bounds below should be read as qualitative guides to the design rather than certificates.

\begin{proposition}[One Euler step]
\label{prop:onestep}
For any $\tau_0\in[0,1)$ and any $\A$, a single Euler step of the exact field from $\tau_0$ to $1$ returns the posterior mean
\begin{equation}
    \A+(1-\tau_0)\,u(\A,\tau_0,\bz)=\E\!\left[\A^1\,\middle|\,\A^{\tau_0}=\A,\bz\right].
\end{equation}
In particular, a cold start ($\tau_0=0$) returns $\E[\A^1\mid\bz]$ for every noise sample, i.e.\ the mean-squared-error behaviour-cloning solution.
\end{proposition}
\begin{pf}
From~\eqref{eq:interp}, $\A^0=(\A^{\tau_0}-\tau_0\A^1)/(1-\tau_0)$, hence $\A^1-\A^0=(\A^1-\A^{\tau_0})/(1-\tau_0)$. Taking conditional expectations given $\A^{\tau_0}=\A$ yields $u(\A,\tau_0,\bz)=(\E[\A^1\mid\A^{\tau_0}=\A,\bz]-\A)/(1-\tau_0)$, which rearranges to the claim. For $\tau_0=0$, $\A^0=\beps$ is independent of $\A^1$ given $\bz$, so the conditioning on $\A^0$ drops out.
\end{pf}

Proposition~\ref{prop:onestep} is elementary; the tendency of large flow steps to average plausible targets is also noted in~\cite{frans2025shortcut}. It has two practical consequences for locomotion. First, a one-step cold-start flow policy averages modes exactly as behaviour cloning does, so representing the multimodal pooled data requires $N\geq2$. Second, a one-step \emph{warm-started} update returns the mean of the posterior $p(\A^1\mid\A^{\tau_0},\bz)\propto p(\A^1\mid\bz)\,\N\big(\A^{\tau_0};\tau_0\A^1,(1-\tau_0)^2I\big)$. As $\tau_0$ grows, this posterior concentrates on the mode containing the previous plan. Warm-starting thus replaces mode \emph{averaging} by mode \emph{commitment}, which is what makes very small NFE budgets viable.

\begin{proposition}[Lipschitz feedback law]
\label{prop:lipschitz}
Let Assumption~\ref{ass:lip} hold, let $\Psi(\A;\bz)$ denote Euler integration~\eqref{eq:euler} of $\vt$ from $\tau_s=n_s/N$ to $1$ starting at $\A$, and let $\rho=(1+L_A/N)^{N-n_s}$. Then $\rho\leq e^{L_A(1-\tau_s)}$ and, for all $\A,\A',\bz,\bz'$,
\begin{multline}
    \norm{\Psi(\A;\bz)-\Psi(\A';\bz')}\\
    \leq\rho\norm{\A-\A'}+\frac{L_z(\rho-1)}{L_A}\norm{\bz-\bz'}.
    \label{eq:lip}
\end{multline}
Consequently, the frozen-noise cold-start policy $\pi(\Ob)=[\Psi(\beps;\M(\Ob))]_1$ is Lipschitz in $\Ob$ with constant $L_\pi=L_ML_z(e^{L_A}-1)/L_A$.
\end{proposition}
\begin{pf}
Let $e_n$ be the distance between the two Euler iterates after $n$ steps and $\delta=\norm{\bz-\bz'}$. By the triangle inequality and Assumption~\ref{ass:lip}, $e_{n+1}\leq(1+L_A/N)e_n+(L_z/N)\delta$. Unrolling over $N-n_s$ steps gives $e_{N-n_s}\leq\rho\,e_0+(L_z\delta/N)\sum_{j<N-n_s}(1+L_A/N)^j=\rho\,e_0+L_z\delta(\rho-1)/L_A$, and $(1+x/N)^m\leq e^{xm/N}$. The corollary follows with $e_0=0$, $\tau_s=0$, the Lipschitz encoder, and the fact that extracting the first action is $1$-Lipschitz.
\end{pf}

With frozen noise, the change in action between consecutive control steps is therefore driven only by the change in observations, $\norm{\hat\ba_k-\hat\ba_{k-1}}\leq L_\pi\norm{\Ob_k-\Ob_{k-1}}$. With warm-starting, \eqref{eq:lip} applies with $\norm{\A-\A'}=\tau_0\norm{\Atil_k-\Atil_k'}$, so a perturbation of the previous plan is scaled by at most $\tau_0\rho$ and is contracted whenever $\tau_0\rho<1$. Samplers that draw fresh noise behave very differently.

\begin{remark}[Jitter floor of fresh-noise sampling]
\label{rem:jitter}
Suppose that a policy draws independent noise at every control step, so that $\ba_k\sim p_\theta(\cdot\mid\Ob_k)$ independently across $k$. This covers ancestral DDPM sampling as in \birodiff{}, as well as DDIM or flow sampling with resampled initial noise. If $\Ob_k=\Ob_{k-1}=\Ob$, then $\ba_k$ and $\ba_{k-1}$ are i.i.d.\ and
\begin{equation}
    \E\norm{\ba_k-\ba_{k-1}}^2=2\,\tr\Cov_{p_\theta}[\ba\mid\Ob].
\end{equation}
This floor is largest exactly where the policy is multimodal, because the covariance then contains the spread between modes. With pooled multi-gait data it corresponds to switching between the experts' gaits mid-stride (Sec.~\ref{sec:intro}).
\end{remark}

\begin{remark}[Gait selection by the noise sample]
\label{rem:select}
With a deterministic sampler, the selected mode is a deterministic function of $\beps$ and the observation history. Freezing $\beps$ per episode therefore removes the noise-driven source of gait switching. Drawing $\beps$ afresh in each episode preserves diversity across episodes, so that over many episodes the policy can reproduce the mixture of expert gaits. Behaviour cloning instead returns the conditional mean (Proposition~\ref{prop:onestep}), a single deterministic action per observation that need not correspond to either gait. Sec.~\ref{sec:protocol} defines metrics for both effects.
\end{remark}

Freezing the noise also removes the floor of Remark~\ref{rem:jitter} for deterministic diffusion samplers such as DDIM. The advantage of the flow formulation is that it remains accurate with a handful of steps. We test this distinction directly by comparing against DDIM sampling of the diffusion network with frozen noise (Sec.~\ref{sec:baselines}).

\begin{proposition}[Warm-start error]
\label{prop:warm}
Fix $\bz_k$ and let $p_1=p(\cdot\mid\bz_k)$ be the target chunk distribution. Let $\Phi$ be the flow map of the exact field $u(\cdot,\cdot,\bz_k)$ from $\tau_0$ to $1$, and assume that $u$ is $L_A$-Lipschitz in $\A$. If $\Atil_k\sim\mu$ and $\beps\sim\N(0,I)$ is independent of $\Atil_k$, the warm-started sample $\Ahat_k=\Phi\big((1-\tau_0)\beps+\tau_0\Atil_k\big)$ satisfies
\begin{equation}
    \Wtwo\big(\mathrm{Law}(\Ahat_k),\,p_1\big)\leq\tau_0\,e^{L_A(1-\tau_0)}\,\Wtwo(\mu,\,p_1).
\end{equation}
\end{proposition}
\begin{pf}
Let $(\Atil_k,\A^1)$ be an optimal coupling of $(\mu,p_1)$, independent of $\beps$, and define $X=(1-\tau_0)\beps+\tau_0\Atil_k$ and $Y=(1-\tau_0)\beps+\tau_0\A^1$. By~\eqref{eq:interp}, $Y$ has the training marginal $p_{\tau_0}(\cdot\mid\bz_k)$, which the exact flow transports to $p_1$, so $\Phi(Y)\sim p_1$. By Gr\"onwall's inequality, $\Phi$ is $e^{L_A(1-\tau_0)}$-Lipschitz. Hence $\Wtwo^2(\mathrm{Law}(\Phi(X)),p_1)\leq\E\norm{\Phi(X)-\Phi(Y)}^2\leq e^{2L_A(1-\tau_0)}\tau_0^2\,\E\norm{\Atil_k-\A^1}^2$, and the last expectation equals $\Wtwo^2(\mu,p_1)$.
\end{pf}

The error of warm-starting is thus proportional to how implausible the shifted previous plan is under the \emph{current} observation. At a control rate of $50$\,Hz, consecutive observations differ only slightly and the shifted plan is a good proposal, except for its duplicated last element. When the terrain changes abruptly, $\Wtwo(\mu,p_1)$ grows, and the factor $\tau_0$ determines how quickly the policy abandons its plan. Proposition~\ref{prop:warm} covers the stochastic variant. The frozen-noise default trades this distributional guarantee for the determinism of Proposition~\ref{prop:lipschitz}.

%% file: figures/arch.tex
\begin{tikzpicture}[
  font=\footnotesize,
  >={Latex[length=1.8mm]},
  box/.style={draw, rounded corners=2pt, align=center, inner sep=3pt, minimum height=9mm},
  dat/.style={box, fill=gray!8},
  enc/.style={box, fill=blue!10},
  net/.style={box, fill=orange!15},
  ctl/.style={box, fill=green!10},
  arr/.style={->, thick},
  lab/.style={font=\scriptsize, inner sep=1pt},
]
\node[dat, text width=19mm] (obs) at (0,0) {Proprioceptive\\ history $\Ob_k$\\ $\in\R^{225}$ ($5{\times}45$)};
\node[enc, text width=14mm] (enc) at (2.55,0) {Encoder $\M$\\ MLP $3{\times}256$};
\node[dat, text width=9mm] (z) at (4.45,0) {$\bz_k$\\ $\in\R^{128}$};
\node[net, text width=23mm] (vel) at (7.0,0) {Velocity field\\ $\vt(\A^\tau,\tau,\bz_k)$\\ MLP $7{\times}512$};
\node[net, text width=21mm] (eul) at (10.15,0) {Euler step\\ $\A\leftarrow\A+\Delta\,\vt$};
\node[dat, text width=25mm] (chunk) at (13.35,0) {Action chunk\\ $\Ahat_k=(\hat\ba_k,\dots,\hat\ba_{k+H-1})$};
\node[ctl, text width=15mm] (exec) at (16.2,0) {Execute $\hat\ba_k$\\ (joint PD)};
\node[ctl, text width=19mm] (robot) at (16.2,-2.35) {Unitree G1\\ (MuJoCo)};

\node[box, fill=white, text width=19mm, minimum height=6mm] (temb) at (7.0,-1.45) {$\gamma(\tau)\in\R^{128}$ {\scriptsize(sinusoidal)}};

\node[dat, text width=19mm] (noise) at (3.6,1.85) {$\beps\sim\N(0,I)$\\ {\scriptsize(frozen per episode)}};
\node[net, fill=orange!6, text width=34mm] (init) at (8.55,1.85) {Warm start at flow time $\tau_0$:\\ $\A^{\tau_0}=(1-\tau_0)\,\beps+\tau_0\,\Atil_k$};
\node[dat, text width=25mm] (shift) at (13.35,1.85) {Shift, delay $z^{-1}$\\ $\Atil_{k+1}=\shift(\Ahat_k)$};

\draw[arr] (obs) -- (enc);
\draw[arr] (enc) -- (z);
\draw[arr] (z) -- (vel);
\draw[arr] (vel) -- (eul);
\draw[arr] (eul) -- (chunk);
\draw[arr] (chunk) -- (exec);
\draw[arr] (exec) -- (robot);
\draw[arr] (temb) -- (vel);
\draw[arr] (noise) -- (init);
\draw[arr] (init.south -| vel.north) -- (vel.north);
\draw[arr] (chunk) -- (shift);
\draw[arr] (shift.west) -- (init.east);
\draw[arr] (eul.south) to[out=-100, in=-60, looseness=1.1] (vel.south east);
\node[lab, align=center] at (10.15,-1.45) {repeat $N(1-\tau_0)$ times\\ ($\tau\leftarrow\tau+\Delta$)};
\draw[arr] (robot.west) -- node[lab, above, pos=0.08]{$\bs_{k+1}$} (obs.south |- robot.west) -- (obs.south);

\begin{scope}[on background layer]
  \node[draw, dashed, rounded corners=3pt, fill=orange!3, inner sep=4pt, fit=(vel)(eul)(temb)(init)] (grp) {};
\end{scope}
\node[lab, anchor=south west] at (grp.north west) {\textbf{ODE integration from $\tau_0$ to $1$}};
\end{tikzpicture}

%% file: sections/experiments.tex
\section{Experimental Setup}
\label{sec:experiments}

The experiments answer three questions.
\begin{itemize}
    \item \textbf{Q1:} How much inference cost does \ours{} save?
    \item \textbf{Q2:} Does it walk reliably and transfer to unseen commands, terrain and perturbations?
    \item \textbf{Q3:} Does it reproduce the experts' gaits without mixing them, and with smooth actions, as Proposition~\ref{prop:lipschitz} and Remarks~\ref{rem:jitter} and~\ref{rem:select} suggest?
\end{itemize}

\subsection{Data Collection}
\label{sec:data}
Each expert is recorded on the platform of Sec.~\ref{sec:platform} in episodes of $20$\,s with a random initial heading. A new command is drawn every $5$\,s, uniformly from the training set $\mathcal{C}_{\text{train}}$: $v_x\in[0,0.5]$\,m/s excluding the held-out band $(0.25,0.35)$, $v_y\in[-0.15,0.15]$\,m/s and $\omega_z\in[-0.4,0.4]$\,rad/s. To broaden state coverage, random horizontal pushes of up to $40$\,N lasting $0.15$\,s are applied on average every $3$\,s. We collect $250$ episodes per expert, i.e.\ $250{,}000$ observation--action pairs each and $500{,}000$ pairs in total, the size of the \birodiff{} dataset~\cite{mothish2024birodiff}. No episode ended in a fall. Episodes are split $80/20$ into training and validation sets. All data are recorded on flat ground, and the learner observes neither the gait clock nor the expert label.

\subsection{Baselines}
\label{sec:baselines}
\textbf{Experts.} We report the fast and slow experts for context. They are not upper bounds, since they observe the gait clock, and a distilled policy may interpolate between behaviours.

\textbf{Learned policies.} All learned policies share the encoder $\M$, the data and the training budget, and differ only in the action head and sampler.
\begin{itemize}
    \item \textbf{MLP-BC:} a deterministic $7{\times}512$ MLP head that regresses $\ba_k$ with an MSE loss. By Proposition~\ref{prop:onestep}, it is the limit of a one-step cold-start flow policy.
    \item \textbf{DDPM}~\cite{mothish2024birodiff}: the \birodiff{} formulation with the same head, $60$ denoising steps, the linear $\beta$ schedule of~\cite{ho2020denoising} rescaled to $60$ steps, and ancestral (stochastic) sampling.
    \item \textbf{DDIM-$K$:} the same trained network sampled with deterministic DDIM~\cite{song2021ddim} using $K\in\{10,4\}$ steps and frozen noise. It matches the determinism and NFE budget of \ours{}, so it isolates the contribution of the flow formulation itself.
    \item \textbf{\ours{} ($H{=}1$):} a single-action flow policy with cold-start sampling ($N=4$). It isolates flow matching from chunking and warm-starting.
\end{itemize}
The warm-started chunked variant of Sec.~\ref{sec:warm} ($H=8$, $\tau_0=0.5$) is included in the cost analysis. Its closed-loop evaluation was not completed and is left to future work.

\subsection{Evaluation Protocol and Metrics}
\label{sec:protocol}
Each learned policy is trained with one seed. Each (policy, condition) pair is evaluated on $M=50$ episodes of $20$\,s, and the same $50$ episode seeds (initial heading, terrain and command sequence) are used for every policy. Besides flat ground with commands from $\mathcal{C}_{\text{train}}$ (\emph{seen}), we test the held-out command $v_x=0.3$\,m/s (\emph{interpolation}) and $v_x=0.6$\,m/s (\emph{extrapolation}). We also test six conditions absent from the data: rough terrain (a random height field), a slope, a single step, lateral pushes on the pelvis ($0.2$\,s, every $4$\,s), reduced friction, and an added torso payload. For the slope and step, commands are restricted to forward walking so that the robot reaches the feature. The severity of each condition was calibrated with the experts only, before any learned policy was evaluated. We chose the first level of a sweep at which the better expert succeeds in $50$--$80\%$ of $20$ episodes, which gave a $24$\,cm height-field amplitude, a $33^\circ$ slope, a $5$\,cm step, $170$\,N pushes, friction $\mu=0.1$ and a $16$\,kg payload. We report:
\begin{itemize}
    \item \textbf{Success rate} (SR): the fraction of episodes in which the pelvis never drops below $0.35$\,m.
    \item \textbf{Velocity tracking:} the planar error $e_v=|v_x-c_x|+|v_y-c_y|$, averaged over time from $1.5$\,s after each command change.
    \item \textbf{Action jitter:} $J=\frac{1}{K}\sum_k\norm{\ba_k-\ba_{k-1}}^2$, the quantity bounded in Proposition~\ref{prop:lipschitz} and Remark~\ref{rem:jitter}, together with the mean mechanical power $\bar P=\frac{1}{K}\sum_k\sum_j|\tau_{j,k}\dot q_{j,k}|$.
    \item \textbf{Gait-mode fidelity:} the stride period of every stride, computed from left-foot touchdowns. We report the $1$-Wasserstein distance $W_1^{\text{gait}}$ between the stride-period distribution of a policy and that of the pooled data. We also label each stride fast or slow by thresholding its period at $0.8$\,s and count \emph{mode switches}, i.e.\ label changes between consecutive strides of an episode. Mode switches are the empirical counterpart of Remarks~\ref{rem:jitter} and~\ref{rem:select}.
    \item \textbf{Inference cost:} NFE and floating-point operations per control step, computed exactly from the architecture.
\end{itemize}

\subsection{Implementation Details}
Table~\ref{tab:hparams} lists the hyperparameters. The diffusion baseline and \ours{} share all of them except the action head and the sampler. As anticipated by Proposition~\ref{prop:onestep}, we parameterise the velocity field of \ours{} through a denoiser $D_\theta(\A,\tau,\bz)$ that predicts the clean chunk, $\vt=(D_\theta-\A)/(1-\tau)$, and train $D_\theta$ by regression onto $\A$. The minimiser $D^\star=\E[\A^1\mid\A^\tau,\bz]$ yields exactly the marginal velocity field, but the network no longer has to reproduce its noisy input at the output. With the plain velocity target $\A-\beps$, the $96$-dimensional chunk head trained much more slowly. All models are trained with Adam~\cite{kingma2014adam} for $30{,}000$ steps at a learning rate of $3\cdot10^{-4}$ on the GPU of an Apple M5 Pro, and we keep the checkpoint with the lowest validation loss. Policies run at $f_c=50$\,Hz and send joint-position targets to the PD law~\eqref{eq:pd}.

\begin{table}[t]
\centering
\caption{Hyperparameters of the DDPM baseline (\birodiff{} formulation) and the evaluated \ours{} configuration.}
\label{tab:hparams}
\footnotesize
\setlength{\tabcolsep}{2pt}
\begin{tabular}{lcc}
\toprule
Parameter & DDPM baseline & \ours{} \\
\midrule
Obs.\ / latent dim. & $225$ / $128$ & $225$ / $128$ \\
Encoder $\M$ (depth$\times$width) & $3\times256$ & $3\times256$ \\
Action head (depth$\times$width) & $7\times512$ & $7\times512$ \\
Output & $\ba_k\in\R^{12}$ & $\ba_k\in\R^{12}$ ($H{=}1$) \\
Regression target & noise $\beps$ & clean chunk $\A$ \\
Generative time & $t\in\{1,\dots,60\}$ & $\tau\sim\mathcal{U}[0,1]$ \\
Time embedding & sinusoidal, $128$ & sinusoidal, $128$ \\
Sampler & DDPM ancestral & Euler, $N{=}4$ \\
Warm start / noise & -- / fresh & -- / frozen \\
NFE per control step & $60$ & $4$ \\
Optimiser, learn.\ rate & Adam, $3\cdot10^{-4}$ & Adam, $3\cdot10^{-4}$ \\
Batch size, steps & $4000$, $30$k & $4000$, $30$k \\
\bottomrule
\end{tabular}
\end{table}

%% file: sections/results.tex
\section{Results}
\label{sec:results}

\subsection{Inference Cost (Q1)}
\label{sec:res_cost}
Table~\ref{tab:compute} reports the per-step cost of each policy. The operation counts follow directly from the architecture. The DDPM policy performs $60$ sequential evaluations of a $1.72$\,M-MAC head, i.e.\ $206.4$\,MFLOP per control step, or $10.3$\,GFLOP/s at $50$\,Hz. The evaluated \ours{} configuration ($H{=}1$, $N{=}4$) needs $14.2$\,MFLOP, a $14.6\times$ reduction. Warm-started chunking ($H{=}8$, $\tau_0{=}0.5$) would lower this to $7.65$\,MFLOP ($27\times$). Chunking is almost free: $H=8$ raises the cost per NFE by $5.0\%$ and $H=16$ by $10.7\%$. At equal NFE, \ours{} ($H{=}1$) and DDIM-4 have identical cost, so any difference in their closed-loop behaviour is due to the generative formulation alone. All learned policies are far more expensive than the compact LSTM expert ($0.06$\,MFLOP), which is the price of representing the pooled distribution with a single feed-forward network.

\begin{table}[t]
\centering
\caption{Inference cost per control step. MFLOP $=2\times$ multiply--accumulates of all linear layers, including the encoder. Cold-start variants use $N=4$; the warm-started variant needs $4$ NFE only at $k=0$.}
\label{tab:compute}
\footnotesize
\setlength{\tabcolsep}{4pt}
\begin{tabular}{lccc}
\toprule
Policy & NFE & MFLOP & Params \\
\midrule
RL expert (LSTM) & $1$ & $0.06$ & $0.03$\,M \\
\midrule
MLP-BC & $1$ & $3.73$ & $1.87$\,M \\
DDPM & $60$ & $206.39$ & $1.94$\,M \\
DDIM-10 & $10$ & $34.77$ & $1.94$\,M \\
DDIM-4 & $4$ & $14.17$ & $1.94$\,M \\
\ours{}, $H{=}1$ (evaluated) & $4$ & $14.17$ & $1.94$\,M \\
\ours{}, $H{=}8$, cold & $4$ & $14.86$ & $2.03$\,M \\
\ours{}, $H{=}8$, warm & $2$ & $7.65$ & $2.03$\,M \\
\bottomrule
\end{tabular}
\end{table}

\begin{table*}[t]
\centering
\caption{Success rate (\%) over $50$ episodes of $20$\,s (one training seed). Seen: flat ground with commands from $\mathcal{C}_{\text{train}}$. Interp./Extrap.: flat ground at $v_x=0.3$ and $0.6$\,m/s. The remaining columns are conditions absent from the data, at the severities calibrated on the experts (Sec.~\ref{sec:protocol}); Avg.\ is their mean.}
\label{tab:main}
\footnotesize
\setlength{\tabcolsep}{4.5pt}
\begin{tabular}{l ccc cccccc c}
\toprule
& \multicolumn{3}{c}{Flat ground} & \multicolumn{6}{c}{Unseen conditions} & \\
\cmidrule(lr){2-4}\cmidrule(lr){5-10}
Policy & Seen & Interp. & Extrap. & Rough & Slope & Step & Push & Low $\mu$ & Payload & Avg. \\
\midrule
Fast expert ($T_g{=}0.6$\,s) & 100 & 100 & 100 & 48 & 80 & 10 & 64 & 44 & 84 & 55.0 \\
Slow expert ($T_g{=}1.0$\,s) & 100 & 100 & 100 & 42 & 6 & 50 & 58 & 30 & 8 & 32.3 \\
\midrule
MLP-BC & 82 & 100 & 100 & 0 & 0 & 0 & 0 & 0 & 0 & 0.0 \\
DDPM (60 steps) & 100 & 100 & 100 & 0 & 0 & 0 & 0 & 0 & 0 & 0.0 \\
DDIM-10 & 100 & 100 & 100 & 2 & 2 & 2 & 0 & 0 & 0 & 1.0 \\
DDIM-4 & 100 & 100 & 98 & 0 & 0 & 2 & 0 & 2 & 2 & 1.0 \\
\ours{} ($H{=}1$, $N{=}4$) & 100 & 100 & 100 & 0 & 0 & 0 & 0 & 0 & 0 & 0.0 \\
\bottomrule
\end{tabular}
\end{table*}

\subsection{Reliability and Transfer (Q2)}
\label{sec:res_gen}
Table~\ref{tab:main} reports success rates. On flat ground, all generative policies walk in every episode with seen commands and at the held-out speed $v_x=0.3$\,m/s, and they extrapolate to $v_x=0.6$\,m/s (DDIM-4: $98\%$). MLP-BC falls in $18\%$ of the episodes with seen commands.

Under the six unseen conditions, the experts succeed in $32$--$55\%$ of the episodes on average, by construction of the calibration. In contrast, every learned policy fails in essentially all of them ($0$--$2\%$). This negative result is important. A policy distilled offline from flat-ground data does not inherit the robustness of experts that observe the gait clock and carry recurrent memory, at least at disturbance levels the experts themselves find hard. The robustness that \birodiff{} reported on milder unseen terrain~\cite{mothish2024birodiff} therefore does not automatically carry over to perturbations of this severity. Closing this gap requires data that covers recovery behaviour, which we discuss in Sec.~\ref{sec:conclusion}.

\begin{table}[t]
\centering
\caption{Tracking, smoothness and gait-mode fidelity on flat ground (one seed, $50$ episodes). $e_{v}$: planar velocity error (m/s) for seen commands and the held-out $v_x{=}0.3$\,m/s. $J$: action jitter ($10^{-3}$\,rad$^2$). $W_1^{\text{gait}}$: stride-period distance to the data (s). Sw.: mode switches per episode. Fast: share of episodes dominated by fast strides (data: $50\%$).}
\label{tab:fidelity}
\footnotesize
\setlength{\tabcolsep}{1.9pt}
\begin{tabular}{lccccccc}
\toprule
Policy & $e_v$ seen & $e_v$ held-out & $J$ & $\bar P$ (W) & $W_1^{\text{gait}}$ & Sw. & Fast \\
\midrule
Fast expert & 0.104 & 0.102 & 28.1 & 37.2 & 0.150 & 0.00 & 100\% \\
Slow expert & 0.100 & 0.101 & 17.0 & 24.7 & 0.250 & 0.00 & 0\% \\
\midrule
MLP-BC & 0.134 & \textbf{0.095} & 32.3 & 42.4 & 0.080 & 0.84 & 76\% \\
DDPM (60) & 0.114 & 0.129 & \textbf{20.1} & \textbf{29.1} & 0.047 & 1.38 & 54\% \\
DDIM-10 & 0.115 & 0.131 & 20.5 & 29.6 & \textbf{0.040} & 1.34 & 60\% \\
DDIM-4 & 0.119 & 0.131 & 20.3 & 29.5 & 0.045 & 1.28 & 56\% \\
\ours{} ($H{=}1$) & \textbf{0.104} & \textbf{0.095} & 21.8 & 30.5 & 0.046 & \textbf{0.44} & 42\% \\
\bottomrule
\end{tabular}
\end{table}

\subsection{Gait Mixing, Smoothness and Tracking (Q3)}
\label{sec:res_smooth}
\begin{figure}[t]
    \centering
    \includegraphics[width=\linewidth]{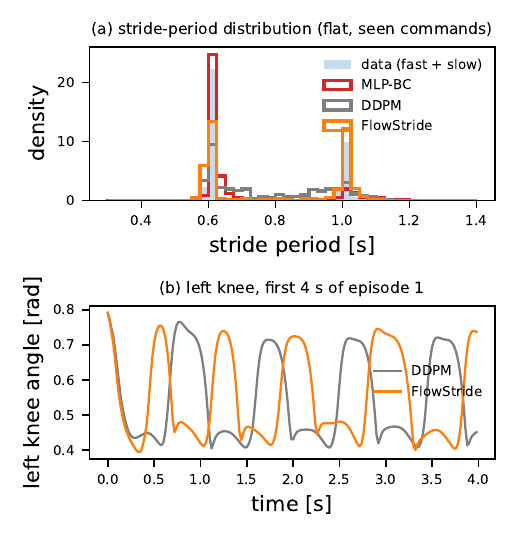}
    \caption{\textbf{Gait-mode fidelity.} (a) Stride-period distribution on flat ground with seen commands ($50$ episodes). The pooled data has two modes at $0.6$ and $1.0$\,s. MLP-BC collapses mostly onto the fast mode, DDPM spreads strides between the modes, and \ours{} reproduces both sharply. (b) Left-knee angle during the first $4$\,s of the same episode for DDPM and \ours{}.}
    \label{fig:gait}
\end{figure}

Fig.~\ref{fig:gait} and Table~\ref{tab:fidelity} examine whether the policies reproduce the two expert gaits without mixing them. MLP-BC does not average the cadences. Instead it largely collapses onto the fast gait: $80\%$ of its strides are shorter than $0.7$\,s and $76\%$ of its episodes are fast-dominated. Its distance to the data distribution ($W_1^{\text{gait}}=0.080$\,s) is $1.7$--$2\times$ that of the generative policies ($0.040$--$0.047$\,s). A unimodal regressor thus loses one of the two behaviours. All generative policies reproduce both modes across episodes, with a share of fast-dominated episodes between $42$ and $60\%$ (data: $50\%$). They differ in how cleanly they do so: $17\%$ of the DDPM strides fall between the modes ($0.7$--$0.9$\,s), against $3\%$ for \ours{}.

They differ markedly in how often they switch gait within an episode. DDPM switches $1.38$ times per episode. \ours{} switches $0.44$ times, a $3.1\times$ reduction, and it does so at the same cost as DDIM-4. DDIM-4 is equally deterministic with frozen noise, yet still switches $1.28$ times, which indicates that the flow formulation, not determinism alone, is responsible for the gait commitment (Remark~\ref{rem:select}). \ours{} also tracks velocity best among the learned policies on seen commands ($e_v=0.104$\,m/s, matching the experts, vs.\ $0.114$ for DDPM and $0.134$ for MLP-BC), and ties MLP-BC at the held-out speed ($0.095$\,m/s vs.\ $0.129$ for DDPM).

Action jitter, in contrast, barely separates the generative policies: $J=20.1$ for DDPM, $20.3$--$20.5$ for DDIM and $21.8\cdot10^{-3}$\,rad$^2$ for \ours{}, all well below MLP-BC ($32.3$). The jitter floor of Remark~\ref{rem:jitter} is proportional to the conditional policy variance, which is evidently small here: the five-step history largely determines the next action, so fresh noise in DDPM adds little high-frequency variation. Where the conditional distribution \emph{is} ambiguous, at gait transitions and after command changes, the different samplers make different discrete choices, which shows up in the mode switches rather than in $J$.

%% file: sections/conclusion.tex
\section{Discussion and Conclusion}
\label{sec:conclusion}

We presented \ours{}, a conditional flow-matching policy for bipedal humanoid locomotion, and studied it on the offline distillation of two gait experts for the Unitree G1. With four Euler steps of a denoiser-parameterised velocity field, \ours{} matched the reliability of a $60$-step DDPM policy on flat ground at $14.6\times$ lower cost. It tracked velocity best among the learned policies and switched gait within an episode $3.1\times$ less often than DDPM, also less often than equally deterministic DDIM sampling at the same cost. Behaviour cloning fell in $18\%$ of the episodes and lost one of the two gaits, confirming that a unimodal regressor is ill-suited to pooled multi-gait data. Our analysis explains these observations: one cold-start step is behaviour cloning, deterministic flow sampling yields a Lipschitz feedback law, and warm-starting has a bounded distributional error.

\textbf{Limitations.} All results are in simulation and use a single training seed. The warm-started chunked variant was analysed and costed but not evaluated in closed loop. Under the severe unseen perturbations used here, which were calibrated so that the experts themselves succeed only partially, every distilled policy failed. Offline imitation from flat-ground data alone does not transfer the experts' robustness. Action jitter did not separate the generative policies, which suggests that the conditional variance given the five-step history is small, except at gait transitions. Our bounds rely on Lipschitz constants that we do not certify.

\textbf{Future work.} The most pressing direction is data that covers recovery behaviour, such as perturbed rollouts or a small amount of on-policy correction, followed by closed-loop evaluation of warm-started chunking over several seeds. Because flow policies admit tractable fine-tuning with RL~\cite{mcallister2025fpo,park2025fql}, \ours{} could also be improved beyond its experts online. Further directions are distillation into a strictly one-step generator~\cite{frans2025shortcut,geng2025meanflow}, on-board deployment on G1 and H1 hardware, and filtering the generated chunks with safety constraints.

%% file: references.bib
@article{ajay2022conditional,
  title={Is conditional generative modeling all you need for decision-making?},
  author={Ajay, Anurag and Du, Yilun and Gupta, Abhi and Tenenbaum, Joshua and Jaakkola, Tommi and Agrawal, Pulkit},
  journal={arXiv preprint arXiv:2211.15657},
  year={2022}
}

@article{gelu,
  title={Gaussian error linear units ({GELUs})},
  author={Hendrycks, Dan and Gimpel, Kevin},
  journal={arXiv preprint arXiv:1606.08415},
  year={2016}
}

@article{ho2020denoising,
  title={Denoising diffusion probabilistic models},
  author={Ho, Jonathan and Jain, Ajay and Abbeel, Pieter},
  journal={Advances in Neural Information Processing Systems},
  volume={33},
  pages={6840--6851},
  year={2020}
}

@article{janner2022planning,
  title={Planning with diffusion for flexible behavior synthesis},
  author={Janner, Michael and Du, Yilun and Tenenbaum, Joshua B and Levine, Sergey},
  journal={arXiv preprint arXiv:2205.09991},
  year={2022}
}

@article{kingma2014adam,
  title={Adam: A method for stochastic optimization},
  author={Kingma, Diederik P and Ba, Jimmy},
  journal={arXiv preprint arXiv:1412.6980},
  year={2014}
}

@article{mandlekar2021matters,
  title={What matters in learning from offline human demonstrations for robot manipulation},
  author={Mandlekar, Ajay and Xu, Danfei and Wong, Josiah and Nasiriany, Soroush and Wang, Chen and Kulkarni, Rohun and Fei-Fei, Li and Savarese, Silvio and Zhu, Yuke and Mart{\'\i}n-Mart{\'\i}n, Roberto},
  journal={arXiv preprint arXiv:2108.03298},
  year={2021}
}

@inproceedings{rudin2022learning,
  title={Learning to walk in minutes using massively parallel deep reinforcement learning},
  author={Rudin, Nikita and Hoeller, David and Reist, Philipp and Hutter, Marco},
  booktitle={Conference on Robot Learning},
  pages={91--100},
  year={2022},
  organization={PMLR}
}

@inproceedings{sohl2015deep,
  title={Deep unsupervised learning using nonequilibrium thermodynamics},
  author={Sohl-Dickstein, Jascha and Weiss, Eric and Maheswaranathan, Niru and Ganguli, Surya},
  booktitle={International Conference on Machine Learning},
  pages={2256--2265},
  year={2015},
  organization={PMLR}
}

@article{levine2020offline,
  title={Offline reinforcement learning: Tutorial, review, and perspectives on open problems},
  author={Levine, Sergey and Kumar, Aviral and Tucker, George and Fu, Justin},
  journal={arXiv preprint arXiv:2005.01643},
  year={2020}
}

@inproceedings{lipman2023flow,
  title     = {Flow Matching for Generative Modeling},
  author    = {Lipman, Yaron and Chen, Ricky T. Q. and Ben-Hamu, Heli and Nickel, Maximilian and Le, Matt},
  booktitle = {International Conference on Learning Representations (ICLR)},
  year      = {2023},
  eprint    = {2210.02747},
  archivePrefix = {arXiv},
}

@inproceedings{liu2023rectified,
  title     = {Flow Straight and Fast: Learning to Generate and Transfer Data with Rectified Flow},
  author    = {Liu, Xingchao and Gong, Chengyue and Liu, Qiang},
  booktitle = {International Conference on Learning Representations (ICLR)},
  year      = {2023},
  eprint    = {2209.03003},
  archivePrefix = {arXiv},
}

@inproceedings{albergo2023building,
  title     = {Building Normalizing Flows with Stochastic Interpolants},
  author    = {Albergo, Michael S. and Vanden-Eijnden, Eric},
  booktitle = {International Conference on Learning Representations (ICLR)},
  year      = {2023},
  eprint    = {2209.15571},
  archivePrefix = {arXiv},
}

@inproceedings{song2021ddim,
  title     = {Denoising Diffusion Implicit Models},
  author    = {Song, Jiaming and Meng, Chenlin and Ermon, Stefano},
  booktitle = {International Conference on Learning Representations (ICLR)},
  year      = {2021},
  eprint    = {2010.02502},
  archivePrefix = {arXiv},
}

@inproceedings{song2021score,
  title     = {Score-Based Generative Modeling through Stochastic Differential Equations},
  author    = {Song, Yang and Sohl-Dickstein, Jascha and Kingma, Diederik P. and Kumar, Abhishek and Ermon, Stefano and Poole, Ben},
  booktitle = {International Conference on Learning Representations (ICLR)},
  year      = {2021},
  eprint    = {2011.13456},
  archivePrefix = {arXiv},
}

@inproceedings{meng2022sdedit,
  title     = {{SDEdit}: Guided Image Synthesis and Editing with Stochastic Differential Equations},
  author    = {Meng, Chenlin and He, Yutong and Song, Yang and Song, Jiaming and Wu, Jiajun and Zhu, Jun-Yan and Ermon, Stefano},
  booktitle = {International Conference on Learning Representations (ICLR)},
  year      = {2022},
  eprint    = {2108.01073},
  archivePrefix = {arXiv},
}

@inproceedings{frans2025shortcut,
  title     = {One Step Diffusion via Shortcut Models},
  author    = {Frans, Kevin and Hafner, Danijar and Levine, Sergey and Abbeel, Pieter},
  booktitle = {International Conference on Learning Representations (ICLR)},
  year      = {2025},
  eprint    = {2410.12557},
  archivePrefix = {arXiv},
}

@inproceedings{geng2025meanflow,
  title     = {Mean Flows for One-step Generative Modeling},
  author    = {Geng, Zhengyang and Deng, Mingyang and Bai, Xingjian and Kolter, J. Zico and He, Kaiming},
  booktitle = {Advances in Neural Information Processing Systems (NeurIPS)},
  volume    = {38},
  year      = {2025},
  eprint    = {2505.13447},
  archivePrefix = {arXiv},
}

@inproceedings{song2023consistency,
  title     = {Consistency Models},
  author    = {Song, Yang and Dhariwal, Prafulla and Chen, Mark and Sutskever, Ilya},
  booktitle = {Proceedings of the 40th International Conference on Machine Learning (ICML)},
  series    = {Proceedings of Machine Learning Research},
  volume    = {202},
  pages     = {32211--32252},
  year      = {2023},
  eprint    = {2303.01469},
  archivePrefix = {arXiv},
}

@inproceedings{black2024pi0,
  title     = {$\pi_0$: A Vision-Language-Action Flow Model for General Robot Control},
  author    = {Black, Kevin and Brown, Noah and Driess, Danny and Esmail, Adnan and Equi, Michael and Finn, Chelsea and Fusai, Niccolo and Groom, Lachy and others},
  booktitle = {Robotics: Science and Systems (RSS)},
  year      = {2025},
  doi       = {10.15607/RSS.2025.XXI.010},
  eprint    = {2410.24164},
  archivePrefix = {arXiv},
}

@inproceedings{braun2024riemannian,
  title     = {Riemannian Flow Matching Policy for Robot Motion Learning},
  author    = {Braun, Max and Jaquier, No{\'e}mie and Rozo, Leonel and Asfour, Tamim},
  booktitle = {IEEE/RSJ International Conference on Intelligent Robots and Systems (IROS)},
  pages     = {5144--5151},
  year      = {2024},
  doi       = {10.1109/IROS58592.2024.10801521},
  eprint    = {2403.10672},
  archivePrefix = {arXiv},
}

@article{zhang2024affordance,
  title     = {Affordance-based Robot Manipulation with Flow Matching},
  author    = {Zhang, Fan and Gienger, Michael},
  journal   = {Frontiers in Robotics and AI},
  volume    = {13},
  pages     = {1906861},
  year      = {2026},
  doi       = {10.3389/frobt.2026.1906861},
  eprint    = {2409.01083},
  archivePrefix = {arXiv},
}

@inproceedings{rouxel2024flow,
  title     = {Flow Matching Imitation Learning for Multi-Support Manipulation},
  author    = {Rouxel, Quentin and Ferrari, Andrea and Ivaldi, Serena and Mouret, Jean-Baptiste},
  booktitle = {IEEE-RAS International Conference on Humanoid Robots (Humanoids)},
  pages     = {528--535},
  year      = {2024},
  doi       = {10.1109/Humanoids58906.2024.10769838},
  eprint    = {2407.12381},
  archivePrefix = {arXiv},
}

@inproceedings{zhang2025flowpolicy,
  title     = {{FlowPolicy}: Enabling Fast and Robust {3D} Flow-Based Policy via Consistency Flow Matching for Robot Manipulation},
  author    = {Zhang, Qinglun and Liu, Zhen and Fan, Haoqiang and Liu, Guanghui and Zeng, Bing and Liu, Shuaicheng},
  booktitle = {Proceedings of the AAAI Conference on Artificial Intelligence},
  volume    = {39},
  number    = {14},
  pages     = {14754--14762},
  year      = {2025},
  doi       = {10.1609/aaai.v39i14.33617},
  eprint    = {2412.04987},
  archivePrefix = {arXiv},
}

@inproceedings{chisari2024pointflowmatch,
  title     = {Learning Robotic Manipulation Policies from Point Clouds with Conditional Flow Matching},
  author    = {Chisari, Eugenio and Heppert, Nick and Argus, Max and Welschehold, Tim and Brox, Thomas and Valada, Abhinav},
  booktitle = {Proceedings of the 8th Conference on Robot Learning (CoRL)},
  series    = {Proceedings of Machine Learning Research},
  volume    = {270},
  pages     = {982--993},
  year      = {2024},
  eprint    = {2409.07343},
  archivePrefix = {arXiv},
}

@inproceedings{jiang2025streaming,
  title     = {Streaming Flow Policy: Simplifying Diffusion/Flow-Matching Policies by Treating Action Trajectories as Flow Trajectories},
  author    = {Jiang, Sunshine and Fang, Xiaolin and Roy, Nicholas and Lozano-P{\'e}rez, Tom{\'a}s and Kaelbling, Leslie Pack and Ancha, Siddharth},
  booktitle = {Proceedings of the 9th Conference on Robot Learning (CoRL)},
  series    = {Proceedings of Machine Learning Research},
  volume    = {305},
  pages     = {238--257},
  year      = {2025},
  eprint    = {2505.21851},
  archivePrefix = {arXiv},
}

@inproceedings{black2025rtc,
  title     = {Real-Time Execution of Action Chunking Flow Policies},
  author    = {Black, Kevin and Galliker, Manuel Y. and Levine, Sergey},
  booktitle = {Advances in Neural Information Processing Systems (NeurIPS)},
  volume    = {38},
  year      = {2025},
  eprint    = {2506.07339},
  archivePrefix = {arXiv},
}

@inproceedings{park2025fql,
  title     = {Flow {Q}-Learning},
  author    = {Park, Seohong and Li, Qiyang and Levine, Sergey},
  booktitle = {Proceedings of the 42nd International Conference on Machine Learning (ICML)},
  series    = {Proceedings of Machine Learning Research},
  volume    = {267},
  pages     = {48104--48127},
  year      = {2025},
  eprint    = {2502.02538},
  archivePrefix = {arXiv},
}

@inproceedings{mcallister2025fpo,
  title     = {Flow Matching Policy Gradients},
  author    = {McAllister, David and Ge, Songwei and Yi, Brent and Kim, Chung Min and Weber, Ethan and Choi, Hongsuk and Feng, Haiwen and Kanazawa, Angjoo},
  booktitle = {International Conference on Learning Representations (ICLR)},
  year      = {2026},
  eprint    = {2507.21053},
  archivePrefix = {arXiv},
}

@inproceedings{chi2023diffusion,
  title     = {Diffusion Policy: Visuomotor Policy Learning via Action Diffusion},
  author    = {Chi, Cheng and Feng, Siyuan and Du, Yilun and Xu, Zhenjia and Cousineau, Eric and Burchfiel, Benjamin and Song, Shuran},
  booktitle = {Robotics: Science and Systems (RSS)},
  year      = {2023},
  doi       = {10.15607/RSS.2023.XIX.026},
  eprint    = {2303.04137},
  archivePrefix = {arXiv},
}

@inproceedings{zhao2023act,
  title     = {Learning Fine-Grained Bimanual Manipulation with Low-Cost Hardware},
  author    = {Zhao, Tony Z. and Kumar, Vikash and Levine, Sergey and Finn, Chelsea},
  booktitle = {Robotics: Science and Systems (RSS)},
  year      = {2023},
  doi       = {10.15607/RSS.2023.XIX.016},
  eprint    = {2304.13705},
  archivePrefix = {arXiv},
}

@inproceedings{liu2025bid,
  title     = {Bidirectional Decoding: Improving Action Chunking via Guided Test-Time Sampling},
  author    = {Liu, Yuejiang and Hamid, Jubayer Ibn and Xie, Annie and Lee, Yoonho and Du, Maximilian and Finn, Chelsea},
  booktitle = {International Conference on Learning Representations (ICLR)},
  year      = {2025},
  eprint    = {2408.17355},
  archivePrefix = {arXiv},
}

@inproceedings{wang2023diffusionql,
  title     = {Diffusion Policies as an Expressive Policy Class for Offline Reinforcement Learning},
  author    = {Wang, Zhendong and Hunt, Jonathan J. and Zhou, Mingyuan},
  booktitle = {International Conference on Learning Representations (ICLR)},
  year      = {2023},
  eprint    = {2208.06193},
  archivePrefix = {arXiv},
}

@inproceedings{huang2024diffuseloco,
  title     = {{DiffuseLoco}: Real-Time Legged Locomotion Control with Diffusion from Offline Datasets},
  author    = {Huang, Xiaoyu and Chi, Yufeng and Wang, Ruofeng and Li, Zhongyu and Peng, Xue Bin and Shao, Sophia and Nikolic, Borivoje and Sreenath, Koushil},
  booktitle = {Proceedings of the 8th Conference on Robot Learning (CoRL)},
  series    = {Proceedings of Machine Learning Research},
  volume    = {270},
  pages     = {1567--1589},
  year      = {2024},
  eprint    = {2404.19264},
  archivePrefix = {arXiv},
}

@inproceedings{mothish2024birodiff,
  title     = {{BiRoDiff}: Diffusion Policies for Bipedal Robot Locomotion on Unseen Terrains},
  author    = {Mothish, {GVS} and Tayal, Manan and Kolathaya, Shishir},
  booktitle = {Tenth Indian Control Conference (ICC)},
  pages     = {385--390},
  year      = {2024},
  publisher = {IEEE},
  doi       = {10.1109/ICC64753.2024.10883743},
  eprint    = {2407.05424},
  archivePrefix = {arXiv},
}

@article{liao2025beyondmimic,
  title     = {{BeyondMimic}: From Motion Tracking to Versatile Humanoid Control via Guided Diffusion},
  author    = {Liao, Qiayuan and Truong, Takara E. and Huang, Xiaoyu and Gao, Yuman and Tevet, Guy and Sreenath, Koushil and Liu, C. Karen},
  journal   = {Science Robotics},
  volume    = {11},
  number    = {117},
  pages     = {eadx8924},
  year      = {2026},
  doi       = {10.1126/scirobotics.adx8924},
  eprint    = {2508.08241},
  archivePrefix = {arXiv},
}

@inproceedings{siekmann2021blind,
  title     = {Blind Bipedal Stair Traversal via Sim-to-Real Reinforcement Learning},
  author    = {Siekmann, Jonah and Green, Kevin and Warila, John and Fern, Alan and Hurst, Jonathan},
  booktitle = {Robotics: Science and Systems (RSS)},
  year      = {2021},
  doi       = {10.15607/RSS.2021.XVII.061},
  eprint    = {2105.08328},
  archivePrefix = {arXiv},
}

@article{radosavovic2024real,
  title     = {Real-World Humanoid Locomotion with Reinforcement Learning},
  author    = {Radosavovic, Ilija and Xiao, Tete and Zhang, Bike and Darrell, Trevor and Malik, Jitendra and Sreenath, Koushil},
  journal   = {Science Robotics},
  volume    = {9},
  number    = {89},
  pages     = {eadi9579},
  year      = {2024},
  doi       = {10.1126/scirobotics.adi9579},
  eprint    = {2303.03381},
  archivePrefix = {arXiv},
}

@article{li2024versatile,
  title     = {Reinforcement Learning for Versatile, Dynamic, and Robust Bipedal Locomotion Control},
  author    = {Li, Zhongyu and Peng, Xue Bin and Abbeel, Pieter and Levine, Sergey and Berseth, Glen and Sreenath, Koushil},
  journal   = {The International Journal of Robotics Research},
  volume    = {44},
  number    = {5},
  pages     = {840--888},
  year      = {2025},
  doi       = {10.1177/02783649241285161},
  eprint    = {2401.16889},
  archivePrefix = {arXiv},
}

@inproceedings{yi2026fpopp,
  title     = {Flow Policy Gradients for Robot Control},
  author    = {Yi, Brent and Choi, Hongsuk and Singh, Himanshu Gaurav and Huang, Xiaoyu and Truong, Takara E. and Sferrazza, Carmelo and Ma, Yi and Duan, Rocky and others},
  booktitle = {Conference on Robot Learning (CoRL)},
  year      = {2026},
  eprint    = {2602.02481},
  archivePrefix = {arXiv},
}

@misc{huang2026rfpo,
  title     = {{RFPO}: Rectified Flow Policy Optimization for Embodied Control},
  author    = {Huang, Ting and Pan, Lisiyu and Wang, Haoyu and Zhang, Zeyu and Qian, Siyuan and Li, Yanjun and Guo, Yandong and Shi, Boxin and Tang, Hao},
  year      = {2026},
  eprint    = {2610.10453},
  archivePrefix = {arXiv},
  primaryClass = {cs.RO},
}

@inproceedings{kurtz2026gpc,
  title     = {Generative Predictive Control: Flow Matching Policies for Dynamic, Difficult-to-Demonstrate Tasks},
  author    = {Kurtz, Vince and Burdick, Joel W.},
  booktitle = {IEEE International Conference on Robotics and Automation (ICRA)},
  pages     = {16552--16559},
  year      = {2026},
  doi       = {10.1109/ICRA57385.2026.11697338},
  eprint    = {2502.13406},
  archivePrefix = {arXiv},
}

@inproceedings{wang2026omnixtreme,
  title     = {{OmniXtreme}: Breaking the Generality Barrier in High-Dynamic Humanoid Control},
  author    = {Wang, Yunshen and Zhu, Shaohang and Zhi, Peiyuan and Li, Yuhan and Li, Jiaxin and Li, Yong-Lu and Xiao, Yuchen and Wang, Xingxing and Jia, Baoxiong and Huang, Siyuan},
  booktitle = {Robotics: Science and Systems (RSS)},
  year      = {2026},
  doi       = {10.15607/RSS.2026.XXII.031},
  eprint    = {2602.23843},
  archivePrefix = {arXiv},
}

@inproceedings{zhang2025reinflow,
  title     = {{ReinFlow}: Fine-tuning Flow Matching Policy with Online Reinforcement Learning},
  author    = {Zhang, Tonghe and Yu, Chao and Su, Sichang and Wang, Yu},
  booktitle = {Advances in Neural Information Processing Systems (NeurIPS)},
  volume    = {38},
  year      = {2025},
  eprint    = {2505.22094},
  archivePrefix = {arXiv},
}

@inproceedings{hoeg2025sdp,
  title     = {Fast Policy Synthesis with Variable Noise Diffusion Models},
  author    = {H{\o}eg, Sigmund H. and Du, Yilun and Egeland, Olav},
  booktitle = {IEEE International Conference on Robotics and Automation (ICRA)},
  pages     = {4821--4828},
  year      = {2025},
  doi       = {10.1109/ICRA55743.2025.11127858},
  eprint    = {2406.04806},
  archivePrefix = {arXiv},
}

@inproceedings{liu2026legato,
  title     = {Learning Native Continuation for Action Chunking Flow Policies},
  author    = {Liu, Yufeng and Yu, Hang and Zhao, Juntu and Li, Bocheng and Zhang, Di and Li, Mingzhu and Wu, Wenxuan and Hu, Yingdong and others},
  booktitle = {Robotics: Science and Systems (RSS)},
  year      = {2026},
  doi       = {10.15607/RSS.2026.XXII.058},
  eprint    = {2602.12978},
  archivePrefix = {arXiv},
}

@inproceedings{zakka2025playground,
  title     = {Demonstrating {MuJoCo} Playground},
  author    = {Zakka, Kevin and Tabanpour, Baruch and Liao, Qiayuan and Haiderbhai, Mustafa and Holt, Samuel and Luo, Jing Yuan and Allshire, Arthur and Frey, Erik and Sreenath, Koushil and Kahrs, Lueder Alexander and Sferrazza, Carmelo and Tassa, Yuval and Abbeel, Pieter},
  booktitle = {Robotics: Science and Systems (RSS)},
  year      = {2025},
  doi       = {10.15607/RSS.2025.XXI.020},
  eprint    = {2502.08844},
  archivePrefix = {arXiv},
}

@inproceedings{todorov2012mujoco,
  title     = {{MuJoCo}: A Physics Engine for Model-Based Control},
  author    = {Todorov, Emanuel and Erez, Tom and Tassa, Yuval},
  booktitle = {IEEE/RSJ International Conference on Intelligent Robots and Systems (IROS)},
  pages     = {5026--5033},
  year      = {2012},
  doi       = {10.1109/IROS.2012.6386109}
}

@misc{gu2024humanoidgym,
  title     = {Humanoid-Gym: Reinforcement Learning for Humanoid Robot with Zero-Shot {Sim2Real} Transfer},
  author    = {Gu, Xinyang and Wang, Yen-Jen and Chen, Jianyu},
  year      = {2024},
  eprint    = {2404.05695},
  archivePrefix = {arXiv},
  primaryClass = {cs.RO},
}

@inproceedings{dicarlo2018convex,
  title     = {Dynamic Locomotion in the {MIT} Cheetah 3 Through Convex Model-Predictive Control},
  author    = {Di Carlo, Jared and Wensing, Patrick M. and Katz, Benjamin and Bledt, Gerardo and Kim, Sangbae},
  booktitle = {IEEE/RSJ International Conference on Intelligent Robots and Systems (IROS)},
  pages     = {1--9},
  year      = {2018},
  doi       = {10.1109/IROS.2018.8594448}
}

@inproceedings{pratt2006capture,
  title     = {Capture Point: A Step toward Humanoid Push Recovery},
  author    = {Pratt, Jerry and Carff, John and Drakunov, Sergey and Goswami, Ambarish},
  booktitle = {IEEE-RAS International Conference on Humanoid Robots (Humanoids)},
  pages     = {200--207},
  year      = {2006},
  doi       = {10.1109/ICHR.2006.321385}
}

@inproceedings{kajita2001lip,
  title     = {The {3D} Linear Inverted Pendulum Mode: A Simple Modeling for a Biped Walking Pattern Generation},
  author    = {Kajita, Shuuji and Kanehiro, Fumio and Kaneko, Kenji and Yokoi, Kazuhito and Hirukawa, Hirohisa},
  booktitle = {IEEE/RSJ International Conference on Intelligent Robots and Systems (IROS)},
  volume    = {1},
  pages     = {239--246},
  year      = {2001},
  doi       = {10.1109/IROS.2001.973365}
}

@article{carius2020mpcnet,
  title     = {{MPC-Net}: A First Principles Guided Policy Search},
  author    = {Carius, Jan and Farshidian, Farbod and Hutter, Marco},
  journal   = {IEEE Robotics and Automation Letters},
  volume    = {5},
  number    = {2},
  pages     = {2897--2904},
  year      = {2020},
  doi       = {10.1109/LRA.2020.2974653},
  eprint    = {1909.05197},
  archivePrefix = {arXiv},
}

@inproceedings{levine2013gps,
  title     = {Guided Policy Search},
  author    = {Levine, Sergey and Koltun, Vladlen},
  booktitle = {Proceedings of the 30th International Conference on Machine Learning (ICML)},
  series    = {Proceedings of Machine Learning Research},
  volume    = {28},
  number    = {3},
  pages     = {1--9},
  year      = {2013}
}

@misc{unitree2023rlgym,
  author       = {{Unitree Robotics}},
  title        = {Unitree {RL} {GYM}: Reinforcement learning implementation for {Unitree} robots},
  howpublished = {\url{https://github.com/unitreerobotics/unitree_rl_gym}},
  year         = {2023},
  note         = {Pretrained G1 locomotion policy and MuJoCo deployment code}
}

@inproceedings{tayal2026bridging,
  title     = {Bridging Safety and Performance in Offline Reinforcement Learning via Epigraph-Guided Flow Matching},
  author    = {Tayal, Manan and Tayal, Mumuksh and Prakash, Ravi},
  booktitle = {The Fortieth Annual Conference on Neural Information Processing Systems (NeurIPS)},
  year      = {2026}
}

@inproceedings{tayal2026safeflowq,
  title     = {Safe Flow {Q}-Learning: Offline Safe Reinforcement Learning with Reachability-Based Flow Policies},
  author    = {Tayal, M. and Tayal, M. and Prakash, R.},
  booktitle = {Reinforcement Learning Conference (RLC)},
  year      = {2026}
}

@IEEEtranBSTCTL{IEEEexample:BSTcontrol,
  CTLdash_repeated_names = "no"
}

@inproceedings{rao2026reachability,
  title     = {Reachability-Guided Safe Policy Learning for Legged Locomotion},
  author    = {Rao, V Abhijith Narayan and Kolathaya, Shishir and Tayal, Manan},
  booktitle = {RSS 2026 Workshop: From Perception to Action},
  year      = {2026}
}
